\documentclass[runningheads]{llncs}

\usepackage{eccv}

\usepackage{eccvabbrv}

\usepackage{graphicx}
\usepackage{booktabs}

\usepackage[accsupp]{axessibility}  

\usepackage{algorithm}
\usepackage{algpseudocode}
\usepackage{amssymb, amsfonts}
\usepackage{amsmath}
\usepackage{tensor}

\usepackage{multirow}
\usepackage{makecell}
\usepackage{color}
\usepackage{pifont} 
\usepackage{bbding}  
\usepackage[bold]{hhtensor}
\usepackage[misc]{ifsym}

\usepackage{wrapfig}
\usepackage[dvipsnames]{xcolor}

\newcommand{\squishlist}{
 \begin{list}{$\bullet$}
  { \setlength{\itemsep}{0pt}
     \setlength{\parsep}{1pt}
     \setlength{\topsep}{1pt}
     \setlength{\partopsep}{0pt}
     \setlength{\leftmargin}{1.5em}
     \setlength{\labelwidth}{1em}
     \setlength{\labelsep}{0.5em} } }
\newcommand{\squishend}{
  \end{list}  }

\usepackage{hyperref}

\usepackage{orcidlink}

\usepackage{graphicx}
\usepackage[accsupp]{axessibility}  
\usepackage{amsmath}
\usepackage{algorithmicx}
\usepackage{algpseudocode}
\begin{document}

\title{UniScale: Arbitrary-Scale Industrial Anomaly Generation}

\titlerunning{UniScale}

\author{Shilei Zeng\textsuperscript{$\ast$} \and
Linxin Guan\textsuperscript{$\ast$} \and
Xurui Li \and
Yaohan Tang \and
Yu Zhou\textsuperscript{\Letter}}

\renewcommand{\thefootnote}{$\ast$}
\footnotetext[1]{Contributed Equally}

\authorrunning{S.~Zeng et al.}

\institute{School of Electronic Information and Communications,\\
Huazhong University of Science and Technology\\
\email{\{shlzeng,lxguan,xrli\_plus,yhtang\_,yuzhou\}@hust.edu.cn}}

\maketitle
\begin{abstract}
Industrial anomaly inspection faces a major challenge due to the lack of real-world anomaly samples. While generative models are used to create anomaly data, existing methods still struggle when handling small-scale anomalies.
This failure occurs because extreme downsampling in diffusion models causes the information of small anomalies to be lost in the latent space.
To address this, we introduce UniScale, a unified training and inference framework for high-fidelity industrial anomaly generation across arbitrary scales.
During training, we introduce an Error-Suppressed Multi-Scale Training (EMT) strategy, 
which enables the model to learn the rich location-aware textures of anomalies, while suppressing upsampling-induced interpolation errors in texture acquisition, 
ensuring the model is capable of learning small-scale anomalies, while remaining effective for regular scale anomalies.
For inference, we propose Generation-then-Fusion Denoising. 
It decouples anomaly generation from background integration, preventing small anomalies from being overwhelmed.
Extensive experiments demonstrate that our method outperforms state-of-the-art competitors in both anomaly generation quality and downstream detection performance.
It achieves a relative IS(a) improvement of 45.86\% (from 1.81 to 2.64) on VisA and 37.70\% (from 1.22 to 1.68) on MVTec AD 2, 
while also improving the downstream pixel-level IoU by 4.22\% on VisA and AUROC by 6.55\% on MVTec AD 2. Code is available at \href{https://github.com/HUST-SLOW/UniScale}{https://github.com/HUST-SLOW/UniScale}.

\end{abstract}

\section{Introduction}
\label{sec:intro}

\begin{figure}[t]
  \centering
  \includegraphics[width=1\linewidth]{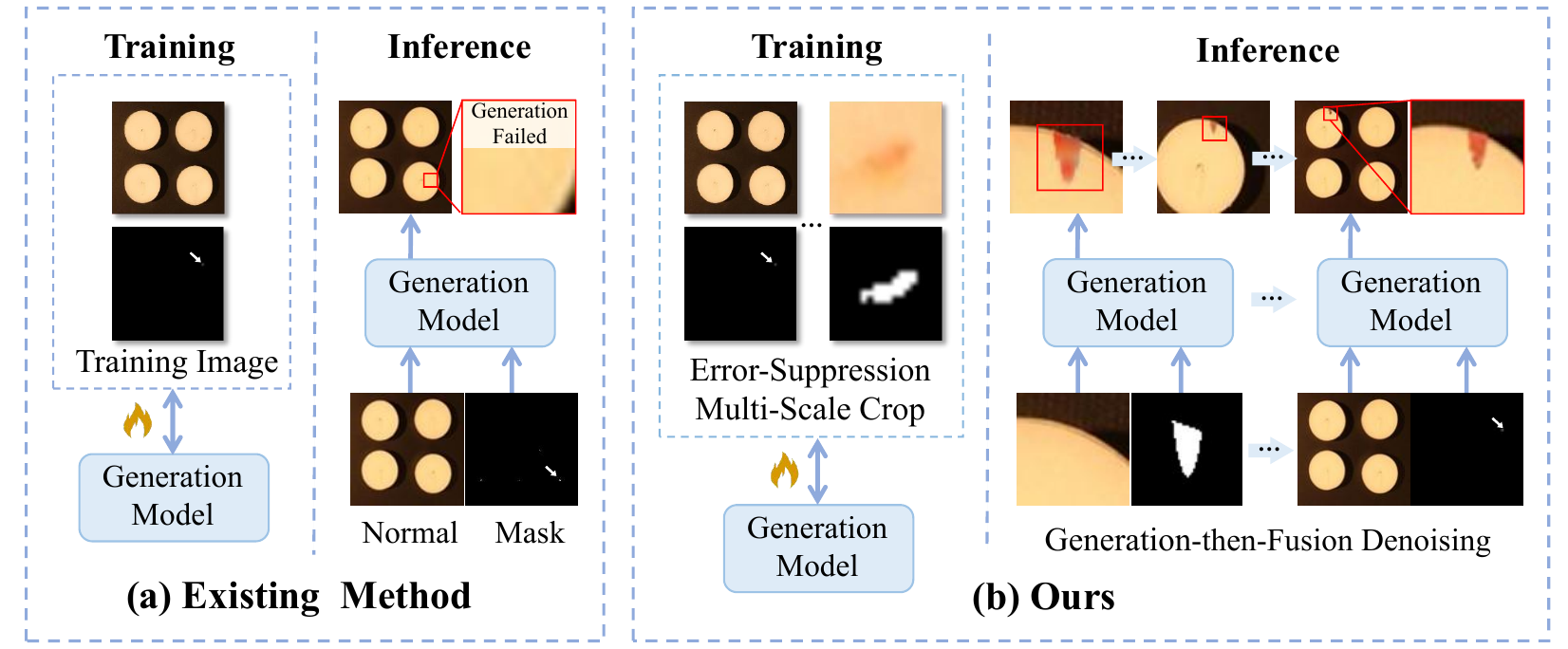}
  \caption{(a) Existing anomaly generation methods often fail to generate small-scale anomalies. In contrast, our method (b) proposes an Error-Suppressed Multi-Scale Training Strategy and Generation-then-Fusion Denoising, successfully generating small-scale anomalies.}
  \label{fig:introduction}
  \vspace{-10pt}
\end{figure}

Industrial anomaly inspection plays a vital role in modern manufacturing, yet anomalous samples are inherently scarce in real-world scenarios. 
This scarcity severely limits the development of reliable inspection systems, as collecting sufficient anomaly data is both challenging and costly. 
To alleviate data insufficiency, recent generative approaches such as AnomalyDiffusion~\cite{hu2024anomalydiffusion}, SeaS~\cite{dai2025seas}, and DualAnoDiff~\cite{jin2025dual} have shown promise in generating anomaly data. 

However, despite recent progress, these methods struggle to handle \textbf{small-scale anomalies}.
Given the typical operating resolution in anomaly generation tasks (e.g., $512 \times 512$), we define anomalies smaller than $32 \times 32$ pixels as small-scale. 
Our statistical analysis shows that such anomalies are very common, accounting for 56.33\% of anomaly images in the VisA dataset. 
The main challenge lies in the extreme downsampling required by the overall model architecture. Specifically, this information loss is a compounded effect caused by both the VAE, which initiates an $8\times$ downsampling, and the U-Net denoising network, which introduces further spatial compression.
As noted in Prompt-to-Prompt~\cite{hertz2022prompt}, the semantics of text-to-image models are mainly controlled by attention maps, which are most effective at low resolutions (e.g., $16 \times 16$). 
This results in a significant $32\times$ downsampling factor compared to the original input. 
Consequently, the information carried by small-scale anomalies is severely compressed or completely lost in the latent space,
making it more challenging for the model to learn these fine details.
Because of this severe information loss, existing methods trained directly on original full-resolution images consistently fail to generate these small anomalies, as visually demonstrated in Fig.~\ref{fig:introduction}(a).

To address these challenges, we propose UniScale, a unified training and inference framework for industrial anomaly generation across arbitrary scales, as shown in Fig.\ref{fig:introduction}(b).
In the training stage, we propose an Error-Suppressed Multi-Scale Training (EMT) strategy, 
which contains a Multi-Scale Cropping (MSC) operation and a Joint Scale (JS) loss. 
EMT extracts anomaly features across varying-scale crops,
optimizes them together, enabling the text embeddings to learn the rich textures of anomalies that satisfy location constraints.
In addition, we propose a Cross-Scale Perception (CSP) loss, which is designed to alleviate the interpolation errors caused by upsampling during texture acquisition. 
By enforcing consistency across scales, CSP loss ensures the text embeddings learn authentic anomaly features, resulting in high-quality anomaly learning.
In the inference stage, we propose Generation-then-Fusion Denoising. 
By separating anomaly generation from its fusion with the product, it preserves fine details while achieving seamless integration with the product, resulting in high-quality anomaly generation across scales.
Experiments show that our method significantly outperforms existing anomaly generation methods on the VisA and MVTec AD 2 datasets.
Regarding anomaly generation quality, our approach achieves a remarkable relative IS(a) improvement of 45.86\% (from 1.81 to 2.64) on the VisA dataset and 37.70\% (from 1.22 to 1.68) on the MVTec AD 2 dataset.
Furthermore, downstream segmentation models trained on our generated data exhibit substantial gains over those trained with existing generation methods.
Specifically,
our method improves the pixel-level IoU by 4.22\% on VisA, 
and achieves a significant performance leap on the more challenging MVTec AD 2 dataset, improving the pixel-level AUROC by 6.55\% compared to the second-best method.

In summary, our contributions are threefold:
\begin{itemize}
    \item We introduce a unified framework for high-fidelity industrial anomaly generation across scales. To the best of our knowledge, this is the first approach capable of handling the particularly challenging task of small-scale industrial anomaly generation.
    \item We propose an Error-Suppressed Multi-Scale Training strategy. It jointly learns fine-grained textures and location-aware anomaly appearance, and suppresses the interpolation errors introduced by the upsampling in texture acquisition.
    In addition, we introduce a Generation-then-Fusion Denoising strategy, which decouples the anomaly generation from background integration, and prevents small anomalies from being ignored during inference, ensuring high-fidelity detail preservation.
    \item Our approach achieves superior performance on the VisA and MVTec AD 2 datasets. Notably, it sets new state-of-the-art records in both generation quality (with a relative IS(a) increase of 45.86\% and 37.70\%, respectively) and downstream anomaly detection
    (improving pixel-level IoU by 4.22\% on VisA, and pixel-level AUROC by 6.55\% on MVTec AD 2),
    confirming its effectiveness for industrial inspection.
\end{itemize}
\section{Related Work}
\label{sec:related}

\subsection{Industrial Anomaly Generation}
Early methods \cite{devries2017cutout, li2021cutpaste, zavrtanik2021draem, schluter2022nsa} primarily relied on data augmentation techniques to synthesize pseudo-anomaly images. However, the significant differences between the generated anomalies and the real ones cause low authenticity. In current studies, some approaches \cite{niu2020sdgan, zhang2021defectgan, Duan2023DFMGAN} employ generation models to enrich the anomalies. 
Some approaches \cite{li2024vague,zhang2024realnet} leverage diffusion models to empower reconstruction-based anomaly detection.
The DFMGAN \cite{Duan2023DFMGAN} is based on StyleGAN2 \cite{karras2020analyzing}, maintaining better structural integrity of the product while generating high-quality anomaly images. Nevertheless, due to insufficient training samples, this method is prone to overfitting.
Further, some methods \cite{hu2024anomalydiffusion, anogen2024, dai2025seas,jin2025dual} achieve anomaly generation based on diffusion models.
SeaS \cite{dai2025seas} and DualAnoDiff \cite{jin2025dual} train diffusion models to learn complete anomaly images, which are capable of generating anomaly images and their corresponding annotations.
AnomalyDiffusion \cite{hu2024anomalydiffusion} decouples anomaly region information into appearance features and positional priors to generate anomaly regions. 
However, existing generative approaches remain limited in addressing small-scale anomalies, 
which are common in industrial inspection.
To overcome such a limitation, we propose an Error-Suppressed Multi-Scale Training strategy. This strategy ensures the model learns both the fine details of small anomalies and their location-dependent appearances, while remaining effective for regular-scale anomalies. 
This achieves faithful generation of small-scale anomalies, without compromising the quality of regular-scale anomaly generation.

\subsection{Subject Tuning in Diffusion Models}
Subject tuning has recently emerged as a powerful paradigm in diffusion models~\cite{ho2020denoising,song2021denoising,ho2021classifier}, aiming to achieve precise text–image binding for specific objects or concepts. It focuses on preserving the identity of a given subject, even under novel prompts or diverse contexts. Representative approaches ~\cite{chen2023disenbooth, shi2024instantbooth, fan2024dreambooth++, huang2024learning, zhou2024migc, chen2025customcontrast, zhu2025multibooth} extend earlier personalization techniques, such as Textual Inversion~\cite{gal2022image} and DreamBooth ~\cite{ruiz2023dreambooth}. 
Tuning and generating small-scale subjects remains a persistent challenge for standard diffusion models, as the fine-grained texture details of diminutive subjects are difficult to learn.
To mitigate this, early strategies~\cite{wang2023patchdiffusion,jian2023stable,feng2024instagen} relied on crop-and-magnify paradigms, such as extracting sparse patches \cite{jian2023stable} or utilizing random crops \cite{feng2024instagen}.
However, purely crop-based fine-tuning isolates small targets, disrupting the seamless integration with the background. To address this, recent approaches, such as Text2Traffic~\cite{lv2025text2traffic} and SOEDiff~\cite{soediff2024}, attempt to jointly train diffusion models on both magnified local patches and original images.
Nevertheless, simply combining these scales is insufficient, as magnifying small crops inevitably introduces interpolation errors. Our method employs Cross-Scale Perception (CSP) loss to suppress the error. By enforcing consistency across scales, the CSP loss ensures that the learnable text embeddings are encouraged to learn consistent anomaly appearances across scales. To the best of our knowledge, this critical challenge of small-subject tuning has not yet been addressed in the field of industrial anomaly generation.
\section{Preliminaries}
\label{sec:Preliminaries}

\noindent \textbf{Stable Diffusion} \cite{rombach2022high} is a text-conditioned latent diffusion framework. 
An input image $x_0$ is first encoded into a latent representation $z_{0}$ using a VAE encoder $\mathcal{E}(\cdot)$, 
which downsamples the image by a factor of 8 into a compact latent space to reduce computational cost. 
Then random noise $\epsilon \sim N(0,I)$ is added at discrete timesteps $t$, producing noisy latent states:
$z_{t} = \sqrt{\bar{\alpha}_{t}} z_0 + \sqrt{1-\bar{\alpha}_{t}} \epsilon$.
Meanwhile, a text prompt $P$ is encoded by a CLIP text encoder into a conditional vector $e$, which guides the U-Net denoising network. 
The training objective minimizes the discrepancy between true and predicted noise:
\begin{equation}
L_{\text{SD}} = \mathbb{E}_{z_0=\mathcal{E}(x_0),P,\epsilon,t} \left[ \|\epsilon - \epsilon_{\theta}(z_t, t, e)\|_2^2 \right].
\label{eq:sd}
\end{equation}
Within the U-Net, cross-attention layers enable direct interaction between textual embeddings and image latents, establishing a text-image binding, 
while self-attention layers capture dependencies among image patches to preserve structural coherence. 
This attention-based mechanism ensures semantic alignment between prompts and visual structures during denoising.

\noindent\textbf{Blended Diffusion} \cite{avrahami2022blended} performs local image editing guided by text. 
It iteratively combines generated content with the original image during the denoising process.
At each sampling step, the model blends the text-guided latent prediction $\hat{z}_{t}$ with the latent representation of the original image $z_{t}$. 
This blending is performed within a binary mask $m$ after the forward diffusion process.
This spatial fusion is formulated as,
\begin{equation}
    z _{t} = \hat{z}_{t} \odot m + z_{t} \odot (1 - m)
\label{eq:blend} 
\end{equation}
This process ensures that the edited region conforms to the textual prompt while preserving seamless coherence with unmodified image areas.

\begin{figure*}[!t]
\begin{center}
\includegraphics[width=1\textwidth]{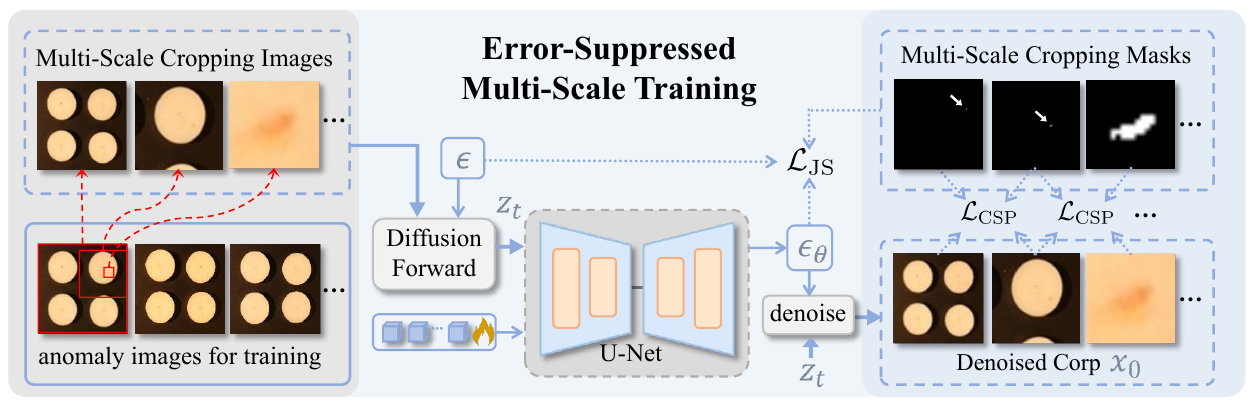}
\caption{
\textbf{Pipeline of Error-Suppressed Multi-Scale Training Strategy}. 
We apply a Multi-Scale Cropping strategy to construct a supervision set.  
Based on this set, we propose a Joint-Scale (JS) loss to bind the anomaly to the text embeddings with a Cross-Scale Perception (CSP) loss to mitigate the interpolation error. 
}
\label{fig:pipeline}
\end{center}
\vspace{-10pt}
\end{figure*}

\section{Method}
\label{sec:method}

We introduce \textbf{UniScale}, a unified training and inference framework for high-fidelity industrial anomaly generation across arbitrary scales.
On the training side, as shown in Fig.~\ref{fig:pipeline}, we introduce the 
Error-Suppressed Multi-Scale Training strategy (Sec.~\ref{sec:cropping}), 
which learns the rich location-aware textures of the anomalies, while suppressing upsampling-induced interpolation errors in texture acquisition.
On the inference side, as shown in Fig.~\ref{fig:pipeline_infer}, we propose Generation-then-Fusion Denoising (GFD) (Sec.~\ref{sec:denoise}), 
which achieves faithful anomaly construction and seamless background integration.

\subsection{Error-Suppressed Multi-Scale Training Strategy}
\label{sec:cropping}

\begin{figure*}[!t]
    \centering
    \includegraphics[width=1\linewidth]{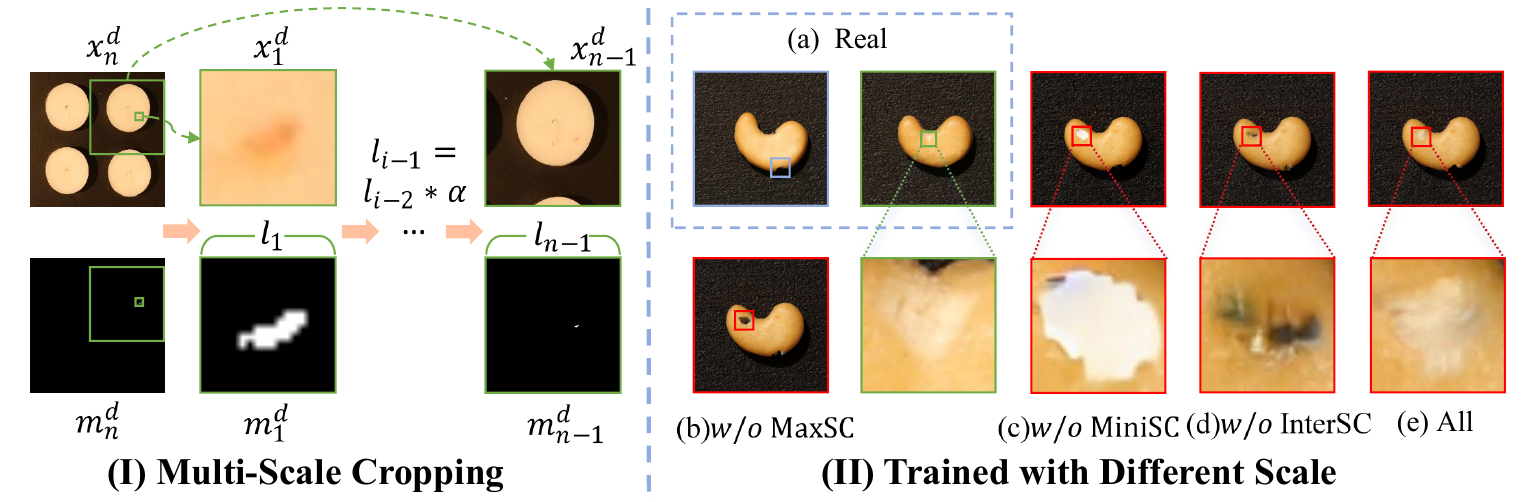}

    \caption{
        \textbf{Multi-Scale Cropping and Training Visualization.}
        (I): Our proposed Multi-Scale Cropping.
        (II): Real VisA dataset images of one anomaly type (blue, green)
        and anomaly generation images trained with different scale crops (red).
    }
    \label{fig:multi_scale}
    \vspace{-5pt}
\end{figure*}

In this section, we first introduce a Multi-Scale Cropping strategy,
shown in Fig.~\ref{fig:multi_scale} (I),
to enable the text embeddings to learn rich textures of the anomalies that satisfy location constraints.
And then we incorporate a Cross-Scale Perception (CSP) loss to alleviate the interpolation error introduced during texture acquisition.

\noindent\textbf{Maximum Scale Crop.}
Given an input RGB anomaly image, we resize it to the generation resolution $\mathcal{S}$ and get $x_n^d \in \mathbb{R}^{\mathcal{S} \times \mathcal{S} \times 3}$ and its corresponding mask $m_n^d \in \mathbb{R}^{\mathcal{S} \times \mathcal{S} \times 3}$,
thereby obtaining the Maximum Scale Crop (MaxSC).
It corresponds to the standard full-image input \cite{hu2024anomalydiffusion,dai2025seas},
and enables the model to learn the location-aware anomaly appearance through the attention mechanism in U-Net, which is critical for correct generation.
As shown in Fig.~\ref{fig:multi_scale}(II)(a), the appearance of breakage depends on its location.
At the edges (blue box), it appears black, which corresponds to the background texture exposed by product damage. 
While inside the product (green box), it shows a whitish texture related to the product itself. 
Without MaxSC, the model tends to learn isolated texture patterns, ignoring the location-aware characteristics of the anomalies, which may lead to incorrect generation.
As illustrated in Fig.~\ref{fig:multi_scale}(II)(b), the model incorrectly generates a black background texture in the center of the product, instead of producing the whitish texture associated with the product.
Therefore, MaxSC is indispensable for preserving the location awareness of anomalies.
The limitation of MaxSC is that, due to the small size of the anomaly region, 
the learnable text embeddings are unable to capture sufficient abnormal texture information during text-image alignment.

\noindent\textbf{Minimum Scale Crop.}
To complement the missing fine-grained texture details of anomalies for text embedding learning,
we further introduce Minimum Scale Crop (MiniSC). 
We first locate the anomaly by extracting the minimum bounding box from the connected components of the mask $m_n^d$. 
Then we extend this bounding box into a square based on its longer side.
This crop is then resized to the training resolution $\mathcal{S}$, 
yielding $x_1^d \in \mathbb{R}^{\mathcal{S} \times \mathcal{S} \times 3}$, its mask $m_1^d \in \mathbb{R}^{\mathcal{S} \times \mathcal{S} \times 3}$, and the corresponding side length $l_1$.
The MiniSC plays a crucial role in preserving anomaly visibility and fine-grained texture details in text embedding learning.
As illustrated in Fig.~\ref{fig:multi_scale}(II)(c), 
when MiniSC is absent, the learned text embeddings tend to guide the generative model to produce blurry regions lacking texture,
which differ significantly from the anomaly textures observed in real data in (II)(a).
However, the problem with MiniSC is that it cannot observe the overall product, making it impossible for U-Net to learn the relationship between the anomalies and the product. This hinders the model from determining whether to generate an anomaly with the correct texture at the correct location. At the same time, since upsampling interpolation alters the original anomaly appearance, it consequently negatively impacts text embedding learning.

\noindent\textbf{Intermediate Scale Crop.}
Although MaxSC and MiniSC are theoretically complementary, 
learning from them alone fails to integrate location constraints with texture information, which may lead to confused generations.
For example, in Fig.~\ref{fig:multi_scale}(II)(d), 
the edge-specific texture and internal anomaly patterns are mixed,
which may reduce the realism of the generated anomalies.
Therefore, we further introduce Intermediate Scale Crop (InterSC), 
by expanding the cropping scope starting from the minimum scale to the maximum scale.
Specifically, let $l_1$ be the side length of the initial minimum bounding box $x_1^d$.
For each subsequent step $i$ ($i > 1$), we enlarge the field of view by increasing the side length: $l_i = l_{i-1} \times \alpha$.
The cropping operation is performed centered on the geometric center of the initial minimum bounding box,
it is formulated as:
\begin{equation}
(x_i^d, m_i^d) = \operatorname{Crop}((x_n^d, m_n^d), l_i), \quad l_i < \mathcal{S},
\label{eq:multi-crop}
\end{equation}
The resulting region is then resized to the training resolution $\mathcal{S} \times \mathcal{S}$, producing a sequence of multi-scale inputs ${(x_i^d, m_i^d)}$.
InterSC progressively reduces the anomaly-to-background ratio in a controlled geometric progression, preserves fine-grained anomaly textures with less interpolation error, and serves as a crucial bridge between MaxSC and MiniSC.
As shown in Fig.~\ref{fig:multi_scale}(II)(e), by using InterSC together with MaxSC and MiniSC, the generation model is enabled to learn how anomaly appearance depends jointly on texture and location.
The choice of $\alpha$ is critical. When $\alpha$ is too large, the transition between scales becomes abrupt, and the intermediate crops fail to bridge texture and positional semantics, resulting in generations similar to those without intermediate crops. Conversely, if $\alpha$ is too small, the number of crops increases excessively, consuming more GPU memory. Therefore, $\alpha$ must be carefully chosen to balance smooth scale transition and GPU memory consumption.

\noindent\textbf{Joint Scale Training.} 
Through the Multi-Scale Cropping process, we construct a multi-scale image set 
$\{(x_1^d,m_1^d),\ldots,(x_n^d,m_n^d)\}$,
ranging from a tightly focused anomaly patch to the full resolution image. 
The number of crops $n$ is adaptively determined by the size of the anomaly.
Since these three scales complement each other, 
learning from them together enables text embedding to learn rich textures that satisfy location constraints.
Therefore, we aggregate the set $\{x_i^d\}_{i=1}^{n}$ into a single training batch, and optimize the following Joint-Scale (JS) loss according to the standard diffusion objective $L_{\text{SD}}$ (Eq.~\ref{eq:sd}):
\begin{equation}
\mathcal{L}_{\text{JS}} =  \sum_{i=1}^{n} \| m^d_i \odot (\epsilon_{i} - \epsilon_{\theta}(z_{i,t}, t, e)) \|_2^2.
\label{eq:JS}
\end{equation}
Here, $z_{i,t}$ represents the noisy latent corresponding to the $i$-th crop $x_i^d$ at timestep $t$, $\epsilon_{i}$ is the target noise, and $e$ is the learnable text embedding shared across all scales.

\begin{figure}[!t]
\centering
\includegraphics[width=0.75\textwidth]{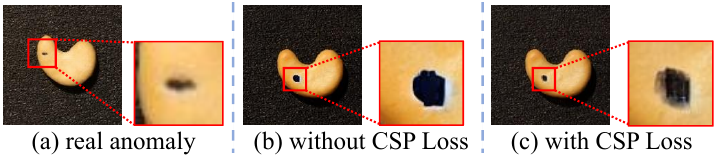}
\caption{
Comparison of generated images w/o CSP loss and w/ (Ours). The model w/o CSP loss cannot generate faithful anomalies.
}
\vspace{-5pt}
\label{fig:lpips}
\end{figure}

\noindent\textbf{Cross-Scale Perception Loss.}
As we mentioned earlier, interpolation  distortions make anomalies appear differently at different scales, causing the text embeddings to learn inconsistent features and produce unrealistic generations. As shown in Fig.~\ref{fig:lpips}(b),  training with JS loss alone produces artifacts that appear unrealistic and lack meaningful texture.
To address this cross-scale inconsistency, we introduce a Cross-Scale Perception (CSP) loss.
The key idea is to enforce perceptual similarity~\cite{zhang2018unreasonable} between anomaly regions at adjacent scales, which are generated under the guidance of the text embeddings. 
In this way, the embeddings are encouraged to learn consistent anomaly features.

During training, for scale $i$, we first predict the noise $\epsilon_{\theta}(z_{i,t}, t, e)$ under the guidance of $e$. 
Then we use the predicted noise to denoise $z_{i,t}$, yielding the clean latent $z_{i,0}$. Then we decode $z_{i,0}$ using the VAE decoder, and get the predicted image $\hat{x}_i^d$, which is used as input for CSP loss.
Then the CSP loss is computed by comparing the perceptual features of $\hat{x}_i^d$ and $\hat{x}_{i+1}^d$ within the anomaly regions. 
Specifically, we crop the anomaly region of higher-resolution scale predicted crops $\hat{x}_{i+1}^d$, use crop side length $l_{i+1}^{\prime} = \mathcal{S} \times l_{i} / l_{i+1}$. 
Then we resize the crop to resolution $\mathcal{S}$, yielding $\hat{x}_{i+1}^{cd}$.
Then we use CSP loss to align it to the lower-resolution scale predicted crops $\hat{x}_i^d$.
The alignment is performed as:
\begin{equation}
\hat{x}_{i+1}^{cd}= \operatorname{Crop}((\hat{x}_{i+1}^d, m_{i+1}^d), l_{i+1}^{\prime})
\end{equation}
where $\hat{x}_{i+1}^d$ is the decoder output at scale $(i+1)$ generated from the latent $\hat{z}_{i+1,0}$. 
Then the CSP loss is defined as:
\begin{equation}
\mathcal{L}_{\text{CSP}} = \sum_{i=1}^{n-1} \mathcal{P}(m_i^d \odot \hat{x}_i^d, m_{i}^d \odot \hat{x}_{i+1}^{cd})
\end{equation}
where $\mathcal{P}$ denotes an off-the-shelf feature extractor used to compute perceptual similarity between anomaly regions at adjacent scales.
By minimizing $\mathcal{L}_{\text{CSP}}$, the learnable text embeddings are encouraged to learn consistent anomaly appearance across scales.
This effectively alleviates the severe interpolation error and the inconsistency introduced by resizing artifacts during multi-scale cropping.

The overall training objective combines the Joint-Scale (JS) loss and the Cross-Scale Perception (CSP) loss:
\begin{equation}
\mathcal{L} = \mathcal{L}_{\text{JS}} + \lambda \cdot \mathcal{L}_{\text{CSP}}.
\end{equation}
The hyperparameter $\lambda$ controls the relative weight of the CSP loss in the final objective.
During training, as in Textual Inversion \cite{gal2022image}, we use learnable tokens whose embeddings are optimized with $\mathcal{L}$.

\subsection{Generation-then-Fusion Denoising}
\label{sec:denoise}
\begin{figure*}[!t]
\begin{center}
\includegraphics[width=1\textwidth]{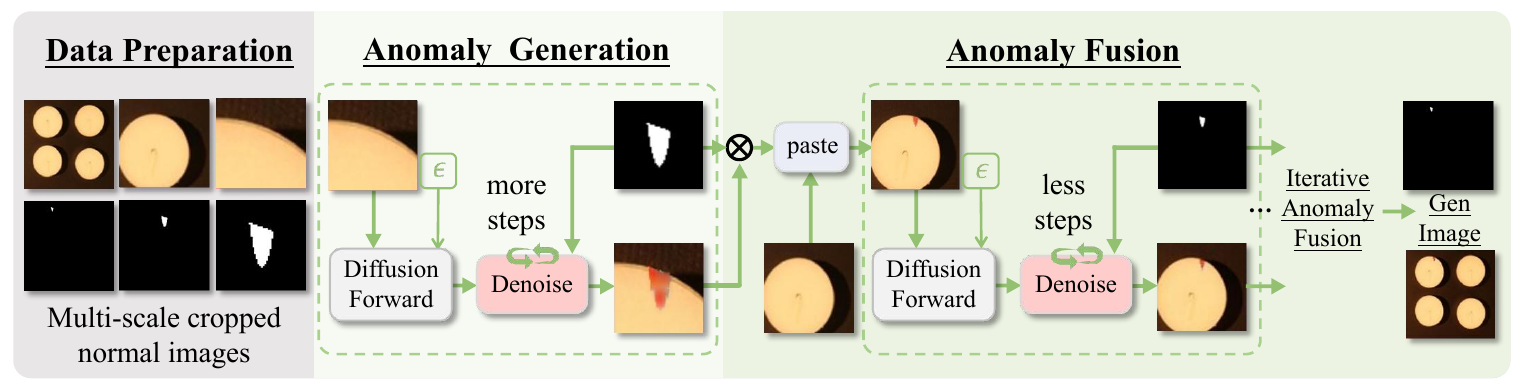}
\caption{
\textbf{Pipeline of the Inference}.
We first crop the normal image, and get the multi-scale normal crops. Then we perform Anomaly Generation and Anomaly Fusion on the crops, and get the final generated anomaly image.
}
\label{fig:pipeline_infer}
\end{center}
\vspace{-20pt}
\end{figure*}

Existing text-to-image inpainting methods typically perform anomaly generation and background fusion within a single denoising stage. 
Due to the low resolution of cross-attention maps, the denoising process is dominated by large-scale background textures and structures. 
When generation and fusion are conducted simultaneously, small anomalies are easily ignored. 
This makes one-stage editing unsuitable for small-scale anomaly generation. 
To overcome this limitation, we introduce Generation-then-Fusion Denoising (GFD), illustrated in Fig.~\ref{fig:pipeline_infer}.

\noindent\textbf{Stage 1: Data Preparation.} 
Given a normal image $x_n^g$, we first generate a stochastic anomaly mask $m_n^g$ using Perlin noise~\cite{ken1985perlin}. The mask $m_n^g$ specifies the spatial regions where anomalies will be injected during subsequent generation. 
Similar to the procedure in Sec.~\ref{sec:cropping}, we extract the minimum bounding box from the connected components of $m_n^g$ and extend it into a square. 
Then we apply the multi-scale cropping strategy (Eq.~\eqref{eq:multi-crop}) to $x_n^g$. Starting from this minimum bounding box, we obtain a set of multi-scale normal crops and masks $\{(x_1^g, m_1^g), \ldots, (x_n^g, m_n^g)\}$, which are used as inputs in the following stages.

\noindent\textbf{Stage 2: Anomaly Generation.}
With the prepared crops and masks, we begin anomaly generation at the minimum crop scale ($i=1$).
The inputs are the normal crop $x_1^g$, the anomaly mask $m_1^g$, and the trained text embeddings $e$.
We generate anomaly content within the masked region by applying $s=50$ steps of Blended Diffusion (Eq.~\eqref{eq:blend}).
The output of this stage is the minimum-scale anomaly image $\hat{x}_1$, which is used as the input for the fusion stage.

\noindent\textbf{Stage 3: Progressive Anomaly Fusion.}
For each subsequent scale ($i > 1$), the inputs are the anomaly image $\hat{x}_{i-1}$ and its mask $m_{i-1}^g$ from the previous scale, together with the current normal crop $x_i^g$ and mask $m_i^g$. 
First, $\hat{x}_{i-1}$ and $m_{i-1}^g$ are downsampled by a factor of $1/\alpha$, producing $\hat{x}_{i-1}^{\downarrow} \in \mathbb{R}^{(\mathcal{S}/\alpha) \times (\mathcal{S}/\alpha) \times 3}$ and $m_{i-1}^{\downarrow} \in \mathbb{R}^{(\mathcal{S}/\alpha) \times (\mathcal{S}/\alpha)}$.
The valid anomaly regions indicated by $m_{i-1}^{\downarrow}$ and the target mask $m_i^g$ are identical in size. 
Next, we use $m_{i-1}^{\downarrow}$ to extract the anomaly pixels from $\hat{x}_{i-1}^{\downarrow}$. 
We then directly paste these pixels into $x_i^g$ at the exact location specified by $m_i^g$. 
This pixel-level replacement is formulated as:
$x_i^g(m_i^g = 1) \leftarrow \hat{x}_{i-1}^{\downarrow}(m_{i-1}^{\downarrow} = 1).$
This operation explicitly injects the anomaly into the higher-resolution crop.
Since direct pasting may cause misalignment with the surrounding background, we perform diffusion forward on the pasted image and perform $s=10$ denoising steps using $m_i^g$ via Eq.~\eqref{eq:blend}, guided by trained text embeddings $e$.
The output of this fusion step is the anomaly image $\hat{x}_i$.
This procedure is repeated iteratively: at each level, the anomaly is injected, aligned, and fused. 
Finally, at the maximum scale ($i=n$), GFD produces the fully anomaly image $\hat{x}_n$.

GFD first constructs a faithful anomaly and then adopts a progressive fusion strategy.
At each scale, the anomaly is reinforced and aligned to the token, 
ensuring that it remains distinguishable while being naturally integrated into the full-resolution image.
\section{Experiments}
\label{sec:experiments}

\subsection{Experimental setting}
\subsubsection{Datasets.}
We conduct our experiments on two established benchmarks, the VisA dataset~\cite{zou2022visa} and MVTec AD 2 dataset~\cite{heckler2025mvtec}. 
VisA dataset consists of 12 objects spanning 3 distinct domains, featuring complex structures, multiple instances, and numerous small-scale anomalies. 
MVTec AD 2 dataset consists of 8 product categories, 
including 2 texture types and 6 object types.
Both datasets contain a substantial proportion of small-scale anomalies (around 50\%), making them well suited for assessing the effectiveness of our arbitrary-scale anomaly generation method.

\begin{table*}[!t]
\scriptsize
\centering
\tabcolsep=4pt
\caption{Comparison on IS, IC-LPIPS (short for IC-L), IS(a) and IC-LPIPS(a) (short for IC-L(a)) on VisA and MVTec AD 2 datasets. Bold denotes the best performance.}
\label{table:generation}
\begin{tabular}{c|cccc|cccc}
\toprule
\multirow{2}{*}{\textbf{Methods}} 
& \multicolumn{4}{c|}{\textbf{VisA}} 
& \multicolumn{4}{c}{\textbf{MVTec AD 2}} \\
\cmidrule(lr){2-5} \cmidrule(lr){6-9}
& \textbf{IS$\uparrow$} & \textbf{IC-L$\uparrow$} & \textbf{IS(a)$\uparrow$} & \textbf{IC-L(a)$\uparrow$} 
& \textbf{IS$\uparrow$} & \textbf{IC-L$\uparrow$} & \textbf{IS(a)$\uparrow$} & \textbf{IC-L(a)$\uparrow$} \\ 
\midrule
Crop\&Paste~\cite{ICME2021crop-and-paste} 
& 1.24 & 0.22 & -- & -- 
& 1.37 & 0.26 & -- & -- \\
DFMGAN~\cite{Duan2023DFMGAN} 
& 1.25 & 0.25 & 1.38 & 0.05 
& 1.39 & 0.29 & 1.04 & 0.05 \\
AnomalyDiffusion~\cite{hu2024anomalydiffusion} 
& 1.26 & 0.25 & 1.33 & 0.04 
& 1.40 & 0.31 & 1.02 & 0.04 \\ 
DualAnoDiff~\cite{jin2025dual} 
& 1.26 & 0.25 & 1.80 & 0.06 
& 1.46 & 0.31 & 1.19 & 0.05 \\ 
SeaS~\cite{dai2025seas} 
& 1.27 & \textbf{0.26} & 1.81 & 0.06 
& 1.42 & 0.32 & 1.22 & 0.05 \\ 
\midrule
\textbf{Ours} 
& \textbf{1.34} & \textbf{0.26} & \textbf{2.64} & \textbf{0.09} 
& \textbf{1.60} & \textbf{0.33} & \textbf{1.68} & \textbf{0.06} \\ 
\bottomrule
\end{tabular}
\end{table*}

\noindent\textbf{Implementation Details.}
We fine-tune the pre-trained Stable Diffusion v1-4~\cite{rombach2022high}. 
We update only the text embedding parameters while keeping all other components frozen. 
For training, we use one-third of the real anomaly images for each anomaly type.
A single unified generative model is trained for each product category to cover all corresponding anomaly types.
The weighting coefficient $\lambda$ is set to 0.001 to ensure that the two loss terms remain on a comparable scale.
The crop expansion factor $\alpha$ is set to 4. 
For each anomaly type, we generate 1,000 synthetic anomaly image-mask pairs to facilitate downstream tasks.
These settings remain consistent across all experiments.

\noindent\textbf{Evaluation Metrics.}
Following SeaS~\cite{dai2025seas}, our evaluation contains two levels: whole images and anomaly regions, using 2 metrics: 
(1) Inception Score (IS)~\cite{salimans2016improved} and Intra-cluster pairwise LPIPS distance (IC-LPIPS) \cite{Ojha_CDC} for authenticity and diversity of whole generated images.
(2) IS and IC-LPIPS calculated only in anomaly regions (denoted as IS(a) and IC-LPIPS(a)) for the authenticity and diversity of anomalies.
For image-level and pixel-level anomaly detection, we use 3 metrics: Area Under Receiver Operating Characteristic curve (AUROC), Average Precision (AP) and F1-score at optimal threshold (F1-max). We also report Intersection over Union (IoU) for segmentation.

\begin{figure}[!t]
\centering
\includegraphics[width=1\textwidth]{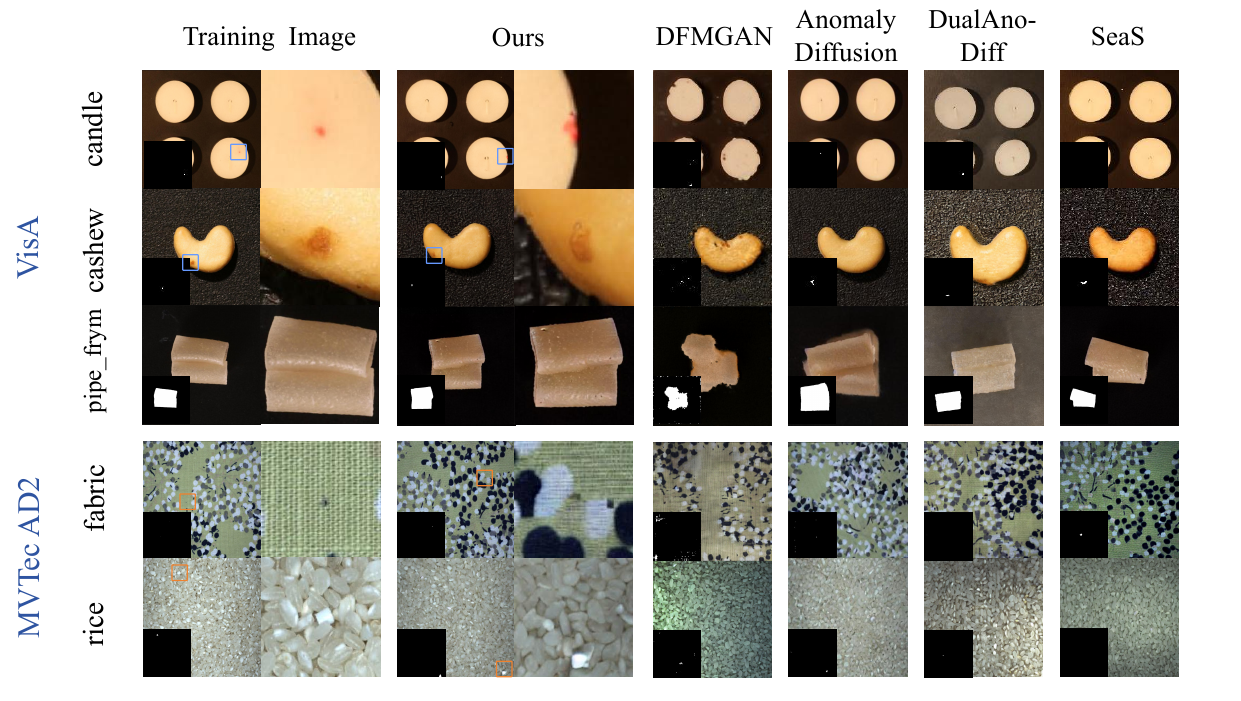}
\caption{Qualitative comparison on VisA and MVTec AD 2. 
Our method generates high-fidelity small-scale anomalies where existing methods fail, while also maintaining structural coherence for larger structural anomalies (e.g., \textit{pipe\_frym}).
}
\label{fig:gen}
\end{figure}

\noindent\textbf{Comparison Methods.}
For the evaluation of anomaly generation quality, we compare our method against current state-of-the-art anomaly generation approaches, including Crop\&Paste~\cite{ICME2021crop-and-paste}, DFMGAN~\cite{Duan2023DFMGAN}, AnomalyDiffusion~\cite{hu2024anomalydiffusion}, DualAnoDiff~\cite{jin2025dual} and SeaS~\cite{dai2025seas}.
To evaluate the generated anomaly image-mask pairs, we train the BiSeNetV2~\cite{yu2021bisenet} segmentation model using the data generated by DFMGAN, AnomalyDiffusion, DualAnoDiff, and SeaS.
In line with the segmentation experimental setup of SeaS, we train a single unified supervised segmentation model across all product categories.
In all comparative experiments, we adopt each method’s original prompt and mask configurations, as their specific prompt designs (e.g., SeaS~\cite{dai2025seas}) and mask acquisition strategies (e.g., DualAnoDiff~\cite{jin2025dual}) constitute their core contributions.

\subsection{Comparison in Anomaly Image Generation}

\noindent\textbf{Quantitative Analysis.}
We compare our method with state-of-the-art anomaly generation methods in terms of fidelity (IS and IS(a)) and diversity (IC-LPIPS and IC-LPIPS(a)).
As presented in Tab.~\ref{table:generation}, our method demonstrates superior performance on both the VisA and MVTec AD 2 datasets.
Specifically, it achieves significant improvements in fidelity metrics, with an average IS(a) increase of 0.83 on VisA (from 1.81 to 2.64) and 0.46 on MVTec AD 2 (from 1.22 to 1.68).
Furthermore, the diversity scores (IC-LPIPS and IC-LPIPS(a)) confirm that our model generates highly diverse anomalies.

\noindent\textbf{Qualitative Analysis.}
We further assess the visual quality of the generated anomaly images.
As shown in Fig.~\ref{fig:gen}, existing methods struggle with small-scale anomalies.
For categories such as \textit{cashew} and \textit{fabric}, existing methods often produce distorted artifacts or fail to generate anomalies altogether.
In contrast, our method generates high-fidelity and realistic anomalies with clear details.
Moreover, our model demonstrates robust generalization capabilities on larger scale structural anomalies (e.g., \textit{pipe\_frym}), successfully generating realistic structural anomalies.

\begin{table*}[!t]
\caption{Comparison of generative models for anomaly detection (Image-level) and anomaly segmentation (Pixel-level). The best-performing result is in bold, the second-best result is underlined.}
\vspace{-10pt}
\label{table:seg}
\renewcommand{\arraystretch}{1.0}
\begin{center}
\setlength{\tabcolsep}{9pt}
\resizebox{1\linewidth}{!}{
\begin{tabular}{l|ccc|cccc}
    \toprule
    \multirow{2}{*}{\makecell{\textbf{Generative}\\\textbf{Models}}} &
    \multicolumn{3}{c|}{\textbf{Image-level}} &
    \multicolumn{4}{c}{\textbf{Pixel-level}} \\
    \cmidrule{2-8}
    & \textbf{AUROC} & \textbf{AP} & \textbf{$F_1$-max} & \textbf{AUROC} & \textbf{AP} & \textbf{$F_1$-max} & \textbf{IoU} \\
    
    \midrule
    \multicolumn{8}{c}{\textbf{VisA}} \\
    \midrule
    DFMGAN & 63.07 & 62.63 & 66.48 & 75.91 & 9.17 & 15.00 & 9.66 \\
    AnomalyDiffusion & 76.11 & 77.74 & 73.13 & 89.29 & 34.16 & 37.93 & 15.93 \\
    DualAnoDiff & 78.06 & 77.92 & 74.76 & 92.06 & 27.12 & 31.94 & 21.48 \\
    SeaS & \underline{85.61} & \underline{86.64} & \underline{80.49} & \underline{96.03} & \underline{42.80} & \underline{45.41} & \underline{25.93} \\
    Ours & \textbf{88.15} & \textbf{88.54} & \textbf{82.01} & \textbf{97.28} & \textbf{42.91} & \textbf{47.03} & \textbf{30.15} \\
    
    \midrule
    \multicolumn{8}{c}{\textbf{MVTec AD 2}} \\
    \midrule
    DFMGAN & 58.01 & 61.22 & 68.91 & 56.31 & 10.12 & 11.53 & 8.22 \\
    AnomalyDiffusion & 61.11 & 63.65 & 71.15 & 59.36 & 13.20 & 15.33 & 9.59 \\
    DualAnoDiff & 63.51 & 69.71 & 74.79 & 64.51 & 14.62 & 16.11 & 11.38 \\
    SeaS & \underline{64.22} & \underline{70.10} & \underline{75.65} & \underline{71.12} & \underline{16.54} & \underline{18.14} & \underline{12.37} \\
    Ours & \textbf{66.26} & \textbf{72.33} & \textbf{76.81} & \textbf{77.67} & \textbf{20.34} & \textbf{24.48} & \textbf{16.06} \\
    \bottomrule
\end{tabular}
}
\end{center}
\vspace{-10pt}
\end{table*}

\subsection{Training supervised downstream models for anomaly segmentation and detection}
We generate 1,000 image-mask pairs for each anomaly type and use them, along with all normal images in the original training sets, to train a unified supervised segmentation model.
The models are tested on the remaining images not included in the training set.

\noindent\textbf{Quantitative Analysis.}
As shown in Tab.\ref{table:seg}, our method outperforms state-of-the-art competitors on both the VisA and MVTec AD 2 datasets.
Specifically, regarding the AUROC metric, our method achieves substantial improvements at both image and pixel levels.
On the VisA dataset, our method sets new state-of-the-art records, reaching 88.15\% image-level AUROC and 97.28\% pixel-level AUROC.
Furthermore, on the challenging MVTec AD 2 dataset, we observe remarkable gains, such as a 6.55\% increase in pixel-level AUROC compared to the second-best method.

\begin{figure}[!t]
\centering
\includegraphics[width=0.98\textwidth]{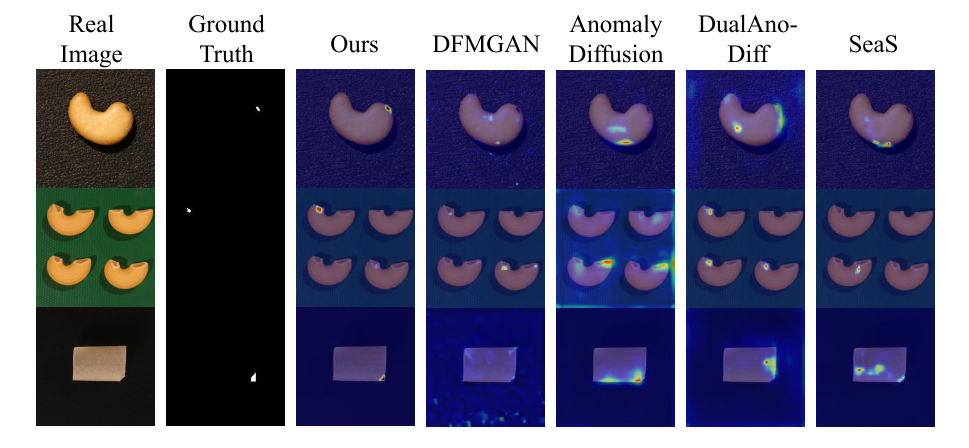}
\caption{
Qualitative results of supervised anomaly segmentation.
}
\label{fig:seg_vis}
\end{figure}

\noindent\textbf{Qualitative Analysis.}
We visually compare the segmentation results of our method against other approaches in Fig.~\ref{fig:seg_vis}.
While existing methods often fail to detect small-scale anomalies, segmentation models trained on our generated image–mask pairs produce precise masks that closely align with the ground truth. 
As illustrated, models trained with our generated image–mask pairs significantly reduce false negatives and ensure high-quality anomaly detection.

\begin{table}[tb]
    \begin{minipage}{0.49\linewidth}
        \caption{Ablation on the core modules of our generation model.}
        \label{tab:ablation_core}
        \centering
        \resizebox{1.0\linewidth}{!}{
            \begin{tabular}{m{2.5cm}|cccccc}
                \specialrule{0.8pt}{0.1pt}{0pt}
                \multirow{2}{*}{\hspace*{1em} Method} & \multicolumn{6}{c}{Metrics} \\
                & IS(a) & IC-L(a) & AUROC & AP & $F_1$-max & IoU \\
                \hline\hline
                (a) w/o EMT & 1.84 & 0.05 & 93.72 & 30.76 & 35.00 & 19.04 \\
                (b) w/o CSP & 2.01 & 0.05 & 97.22 & 31.98 & 37.18 & 27.07 \\
                (c) w/o GFD & 2.08 & 0.05 & 96.03 & 33.62 & 39.51 & 27.97 \\
                \textbf{(d) All (Ours)} & \textbf{2.64} & \textbf{0.09} & \textbf{97.28} & \textbf{42.91} & \textbf{47.03} & \textbf{30.15} \\
                \specialrule{0.8pt}{0.1pt}{0pt}
            \end{tabular}
        }
    \end{minipage}
    \hfill 
    \begin{minipage}{0.49\linewidth}
        \caption{Ablation on Multi-Scale Cropping.}
        \label{tab:ablation_input}
        \centering
        \resizebox{1.0\linewidth}{!}{
            \begin{tabular}{m{2.5cm}|cccccc}
                \specialrule{0.8pt}{0.1pt}{0pt}
                \multirow{2}{*}{Method} & \multicolumn{6}{c}{Metrics} \\
                & IS(a) & IC-L(a) & AUROC & AP & $F_1$-max & IoU \\
                \hline\hline
                w/o MaxSC & 2.58 & 0.07 & 96.03 & 33.62 & 39.51 & 27.97 \\
                w/o MiniSC & 2.04 & 0.05 & 96.43 & 35.21 & 40.59 & 28.72 \\
                w/o InterSC & 2.18 & 0.05 & 93.48 & 34.17 & 39.93 & 29.97 \\
                MaxSC only & 1.84 & 0.05 & 93.72 & 30.76 & 35.00 & 19.04 \\
                MiniSC only & 2.40 & 0.07 & 94.62 & 32.83 & 41.02 & 28.03 \\
                InterSC only & 2.41 & 0.07 & 96.50 & 34.90 & 40.56 & 27.40 \\
                \textbf{Ours} & \textbf{2.64} & \textbf{0.09} & \textbf{97.28} & \textbf{42.91} & \textbf{47.03} & \textbf{30.15} \\
                \specialrule{0.8pt}{0.1pt}{0pt}
            \end{tabular}
        }
    \end{minipage}
\end{table}

\begin{table}[t]
    \begin{minipage}{0.49\linewidth}
        \caption{Ablation on the cropping factor.}
        \label{tab:ablation_alpha}
        \centering
        \resizebox{1.0\linewidth}{!}{
            \begin{tabular}{m{2.5cm}|cccccc|c}
                \specialrule{0.8pt}{0.1pt}{0pt}
                \multirow{2}{*}{Method} & \multicolumn{6}{c|}{Metrics} & \multirow{2}{*}{\begin{tabular}[c]{@{}c@{}}GPU Mem.\\ (GB)\end{tabular}} \\
                & IS(a) & IC-L(a) & AUROC & AP & $F_1$-max & IoU &  \\
                \hline\hline
                $\alpha=2$ & 2.63 & \textbf{0.09} & 96.86 & 41.84 & 45.58 & 30.07 & 24.61 \\
                \textbf{$\alpha=4$ (Ours)} & \textbf{2.64} & \textbf{0.09} & \textbf{97.28} & \textbf{42.91} & \textbf{47.03} & \textbf{30.15} & 20.86 \\
                $\alpha=8$ & 2.55 & 0.08 & 96.21 & 39.50 & 43.43 & 28.62 & 18.91 \\
                $\alpha=16$ & 2.34 & 0.07 & 96.08 & 38.87 & 42.02 & 28.34 & 16.18 \\
                \specialrule{0.8pt}{0.1pt}{0pt}
            \end{tabular}
        }
    \end{minipage}
    \hfill 
    \begin{minipage}{0.49\linewidth}
        \caption{Ablation on GFD.}
        \label{tab:ablation_gfd}
        \centering
        \resizebox{1.0\linewidth}{!}{
            \begin{tabular}{m{3.5cm}|cccccc}
                \specialrule{0.8pt}{0.1pt}{0pt}
                \multirow{2}{*}{Method} & \multicolumn{6}{c}{Metrics} \\
                & IS(a) & IC-L(a) & AUROC & AP & $F_1$-max & IoU \\
                \hline\hline
                (a) One-stage & 2.08 & 0.05 & 96.03 & 33.62 & 39.51 & 27.97 \\
                (b) Gen + paste & 2.52 & 0.08 & 97.02 & 40.40 & 44.60 & 27.65 \\
                (c) Gen + fusion@orig & 2.16 & 0.05 & 96.44 & 34.28 & 40.12 & 28.41 \\
                \textbf{(d) GFD (Ours)} & \textbf{2.64} & \textbf{0.09} & \textbf{97.28} & \textbf{42.91} & \textbf{47.03} & \textbf{30.15} \\
                \specialrule{0.8pt}{0.1pt}{0pt}
            \end{tabular}
        }
    \end{minipage}
\end{table}

\subsection{Ablation Study}

\noindent\textbf{Ablation on core components.}
Tab.\ref{tab:ablation_core} presents the ablation study on the core modules of our framework.
The baseline (a), trained without the Error-Suppressed Multi-Scale Training Strategy (EMT).
In other words, it is trained on original images.
It exhibits the lowest performance, with IS(a) dropping from 2.64 to 1.84 and Pixel-AP decreasing by over 12\%.
This indicates that original-resolution training fails to capture the anomaly details.
However, training with Multi-Scale Cropping alone is insufficient.
Without the CSP loss (b), generation fidelity remains suboptimal (IS(a) 2.01), confirming that suppressing interpolation error is critical for realistic texture generation.
Moreover, replacing our GFD strategy with standard one-stage generation (c) leads to a significant decline in segmentation accuracy (AP 33.62\% vs. 42.91\%).
This demonstrates that our GFD effectively prevents anomalies from being overwhelmed by the background, ensuring high-quality downstream segmentation.

\noindent\textbf{Ablation on Multi-Scale Cropping.}
Tab.\ref{tab:ablation_input} examines the contribution of each scale within our Multi-Scale Cropping strategy (MaxSC, MiniSC and InterSC).
Removing the MaxSC results in the largest AP drop (42.91\% to 33.62\%), indicating that correctly aligning anomaly appearances with their positions is crucial.
Excluding the MiniSC leads to the most significant decrease in IS(a) (2.64 to 2.04), confirming this scale as the primary source of fine-grained textures.
Notably, removing the InterSC causes AUROC to fall from 97.28\% to 93.48\%, suggesting that InterSC is vital for bridging the gap between fine-grained texture details and location-aware anomaly appearance.
Using only InterSC yields suboptimal performance, proving all three scales provide complementary information essential for capturing the dependency between textures and locations.
These results demonstrate that all crops provide complementary information.

\noindent\textbf{Ablation on cropping factor $\alpha$.}
We further study the effect of the cropping factor $\alpha$ in our Multi-Scale Cropping. As shown in Tab.~\ref{tab:ablation_alpha}, a moderate factor ($\alpha=4$) achieves the best overall performance. The choice of $\alpha$ balances smooth scale transitions with GPU memory consumption. If $\alpha$ is too large, the transition between scales becomes abrupt, failing to bridge texture and positional semantics and diluting the anomaly. Conversely, although a smaller $\alpha$ achieves comparable results to $\alpha=4$, it over-crops the anomaly region and increases the number of crops, leading to excessive GPU memory consumption. 
Thus, $\alpha=4$ strikes the optimal balance.

\noindent\textbf{Ablation on Generation-then-Fusion Denoising.}
Tab.\ref{tab:ablation_gfd} evaluates the design choices in our GFD module.
One-stage generation (a) results in low AP as fine details are overwhelmed by the background.
While direct pasting (b) improves fidelity, it yields the lowest IoU (27.65\%) due to severe boundary misalignment.
Performing fusion only at the original scale (c) fails to recover lost details, as seen in the drop in AP compared to our full model.
In contrast, our progressive multi-scale fusion (d) achieves the best performance across all metrics.
These results highlight that decoupling generation from fusion is essential for high-fidelity anomaly generation and accurate downstream localization.

\section{Conclusion}
\label{sec:conclusion}

In this paper, we propose UniScale, a unified training and inference framework for high-fidelity industrial anomaly generation across arbitrary scales. During training, we introduce an Error-Suppressed Multi-Scale Training (EMT) strategy to learn detailed location-aware textures and suppress interpolation errors. Additionally, our Generation-then-Fusion Denoising decouples anomaly generation from background fusion, resulting in high-quality anomaly generation across scales. Extensive experiments on the VisA and MVTec AD 2 datasets validate our method. We establish new state-of-the-art records, achieving relative IS(a) improvements of 45.86\% and 37.70\%.

\section{Acknowledgement}
This work was supported by the National Natural Science Foundation of China under Grant No.62176098. The computation is completed on the HPC Platform of Huazhong University of Science and Technology.
{
    \small
    \bibliographystyle{splncs04}
    \bibliography{main}
}
\appendix
\newpage

\section*{Appendix}                
\setcounter{subsection}{0}         
\renewcommand{\thesubsection}{\Alph{subsection}} 

\subsection{Computational Cost Analysis}
Tab.\ref{tab:cost_benefit} summarizes the computational requirements of UniScale and the baseline methods. UniScale optimizes only the text embeddings while maintaining fixed U-Net parameters, requiring 50 hours of training. During inference, the multi-stage GFD process takes 7 seconds per image, whereas the compared baselines range from 1 to 23 seconds.

\vspace{-15pt}
\begin{table}[h]
\centering
\caption{Computational Cost Analysis.}
\label{tab:cost_benefit}
\setlength{\tabcolsep}{6pt}
\resizebox{1.0\linewidth}{!}{
\begin{tabular}{l|ccc|c}
\toprule
\textbf{Method} & \textbf{Training Memory} & \textbf{Training Time} & \textbf{Inference Time} & \textbf{Pixel-AUROC} \\
\midrule
AnomalyDiffusion~\cite{hu2024anomalydiffusion} & 18 GB & 249 h & 4 s & 89.29 \\
DualAnoDiff~\cite{jin2025dual} & 20 GB & 142 h & 23 s & 92.06 \\
SeaS~\cite{dai2025seas} & 24 GB & 74 h & 1 s & 96.03 \\
\midrule
\textbf{Ours} & 21 GB & 50 h & 7 s & 97.28 \\
\bottomrule
\end{tabular}
}
\end{table}

\vspace{-30pt}
\subsection{Quantitative Evaluation on Small-Scale Anomalies}
To assess performance on small-scale anomalies, we evaluate UniScale on a subset of the VisA dataset containing anomalies smaller than $32\times32$ pixels. Tab.\ref{table:small_scale} reports the comparison against SeaS. 
UniScale improves the generation fidelity (IS(a)) from 1.68 to 2.59 and the diversity (IC-L(a)) from 0.06 to 0.09. Regarding downstream detection tasks, our method yields an image-level AP of 84.07\% (vs. 60.60\% for SeaS) and a pixel-level AP of 31.53\% (vs. 14.03\%). 

\vspace{-15pt}
\begin{table}[h]
\centering
\caption{Results on the small-scale subset ($<32\times32$) of VisA.}
\label{table:small_scale}
\setlength{\tabcolsep}{6pt}
\resizebox{1.0\linewidth}{!}{
\begin{tabular}{l|cc|ccc|ccc}
\toprule
\multirow{2}{*}{\textbf{Method}} & \multicolumn{2}{c|}{\textbf{Generation}} & \multicolumn{3}{c|}{\textbf{Image-level}} & \multicolumn{3}{c}{\textbf{Pixel-level}} \\
& \textbf{IS(a)$\uparrow$} & \textbf{IC-L(a)$\uparrow$} & \textbf{AUROC} & \textbf{AP} & \textbf{$F_1$-max} & \textbf{AUROC} & \textbf{AP} & \textbf{$F_1$-max} \\
\midrule
SeaS~\cite{dai2025seas} & 1.68 & 0.06 & 70.53 & 60.60 & 63.07 & 94.82 & 14.03 & 22.69 \\
\textbf{Ours} & \textbf{2.59} & \textbf{0.09} & \textbf{88.01} & \textbf{84.07} & \textbf{78.50} & \textbf{97.05} & \textbf{31.53} & \textbf{37.75} \\
\bottomrule
\end{tabular}
}
\end{table}

\vspace{-30pt}
\subsection{Results on MPDD Dataset}
To further evaluate the generalization capability of UniScale, we conduct additional experiments on the MPDD dataset. Tab.\ref{table:mpdd} summarizes the quantitative comparison between UniScale and SeaS \cite{dai2025seas}. Regarding generation quality, UniScale achieves an IS(a) of 2.55, outperforming the 1.84 reported by SeaS, which indicates higher fidelity in anomaly synthesis. For downstream anomaly detection, UniScale yields consistent performance gains across both image and pixel levels. Specifically, our method reaches a pixel-level AUROC of 86.68\% and an AP of 39.29\%, compared to 85.18\% and 38.12\% for SeaS, respectively. These results suggest that the proposed multi-scale training and fusion strategies generalize effectively across diverse industrial product categories.

\begin{table}[h]
\centering
\caption{Quantitative evaluation on the MPDD dataset.}
\label{table:mpdd}

\setlength{\tabcolsep}{6pt}
\resizebox{\linewidth}{!}{
\begin{tabular}{l|cc|ccc|ccc}
\toprule
\multirow{2}{*}{\textbf{Method}} & \multicolumn{2}{c|}{\textbf{Generation}} & \multicolumn{3}{c|}{\textbf{Image-level}} & \multicolumn{3}{c}{\textbf{Pixel-level}} \\
& \textbf{IS(a)$\uparrow$} & \textbf{IC-L(a)$\uparrow$} & \textbf{AUROC} & \textbf{AP} & \textbf{$F_1$-max} & \textbf{AUROC} & \textbf{AP} & \textbf{$F_1$-max} \\
\midrule
SeaS~\cite{dai2025seas} & 1.84 & \textbf{0.05} & 80.31 & 83.44 & 83.52 & 85.18 & 38.12 & 40.02 \\
\textbf{Ours} & \textbf{2.55} & \textbf{0.05} & \textbf{81.16} & \textbf{84.21} & \textbf{84.13} & \textbf{86.68} & \textbf{39.29} & \textbf{42.26} \\
\bottomrule
\end{tabular}
}
\end{table}

\subsection{Analysis of Masks}
To determine whether downstream performance gains are attributable to high-fidelity texture synthesis or the utilization of specific mask priors, we conduct a controlled experiment where AnomalyDiffusion \cite{hu2024anomalydiffusion} is trained using the exact same mask configurations as UniScale. As summarized in Tab.\ref{table:mask_prior}, AnomalyDiffusion produces lower detection metrics compared to UniScale, even when provided with identical shape priors. Specifically, our method achieves a pixel-level AUROC of 97.28\% compared to 88.64\% for the baseline.

\vspace{-10pt}
\begin{table}[!t]
\centering
\caption{Comparison using identical mask priors.}
\label{table:mask_prior}
\setlength{\tabcolsep}{6pt}
\resizebox{\linewidth}{!}{
\begin{tabular}{l|cc|ccc|ccc}
\toprule
\multirow{2}{*}{\textbf{Method}} & \multicolumn{2}{c|}{\textbf{Generation}} & \multicolumn{3}{c|}{\textbf{Image-level}} & \multicolumn{3}{c}{\textbf{Pixel-level}} \\
& \textbf{IS(a)$\uparrow$} & \textbf{IC-L(a)$\uparrow$} & \textbf{AUROC} & \textbf{AP} & \textbf{$F_1$-max} & \textbf{AUROC} & \textbf{AP} & \textbf{$F_1$-max} \\
\midrule
AnomalyDiff (Same Mask) & 1.30 & 0.04 & 75.45 & 76.53 & 73.05 & 88.64 & 33.52 & 37.01 \\
\textbf{Ours} & \textbf{2.64} & \textbf{0.09} & \textbf{88.15} & \textbf{88.54} & \textbf{82.01} & \textbf{97.28} & \textbf{42.91} & \textbf{47.03} \\
\bottomrule
\end{tabular}
}
\end{table}

\begin{table}[!t]
\centering
\caption{Comparison of synthesis-based AD methods under different anomaly generation strategies on VisA.}
\label{table:combined_replace}
\setlength{\tabcolsep}{8pt}
\resizebox{\linewidth}{!}{
\begin{tabular}{l|ccc|ccc}
\toprule
\multirow{2}{*}{\textbf{Method}} & \multicolumn{3}{c|}{\textbf{Image-level}} & \multicolumn{3}{c}{\textbf{Pixel-level}} \\
\cmidrule{2-4} \cmidrule{5-7}
& \textbf{AUROC} & \textbf{AP} & \textbf{$F_1$-max} & \textbf{AUROC} & \textbf{AP} & \textbf{$F_1$-max} \\
\midrule
DRAEM \cite{zavrtanik2021draem} & 86.28 & 85.30 & 81.66 & 92.92 & 17.15 & 22.95 \\
DRAEM + SeaS & 88.12 & 87.04 & 83.04 & 98.45 & 49.05 & \textbf{48.62} \\
\textbf{DRAEM + UniScale} & \textbf{91.25} & \textbf{90.63} & \textbf{86.25} & \textbf{98.55} & \textbf{49.26} & 48.51 \\
\midrule
GLASS \cite{chen2024unifiedglass} & 97.68 & 96.89 & 93.03 & \textbf{98.47} & 45.58 & 48.39 \\
GLASS + SeaS & 97.88 & 97.39 & 93.21 & 98.43 & 48.06 & \textbf{49.32} \\
\textbf{GLASS + UniScale} & \textbf{97.88} & \textbf{97.62} & \textbf{93.42} & 98.21 & \textbf{49.22} & 49.21 \\
\bottomrule
\end{tabular}
}
\end{table}

\subsection{Combining Generated Anomalies with Synthesis-Based AD Methods}
To further validate the utility of our generated data, we integrate UniScale into existing synthesis-based anomaly detection (AD) frameworks, such as DRAEM \cite{zavrtanik2021draem} and GLASS \cite{chen2024unifiedglass}. Typically, these methods rely on heuristically synthesized pseudo-anomalies (e.g., Perlin noise combined with external textures), which often lack realism and struggle to represent fine-grained, location-aware defects. 

In this experiment, we replace the standard pseudo-anomalies in these frameworks with anomalies generated by UniScale. As shown in Tab.\ref{table:combined_replace}, incorporating our generated anomalies generally enhances the performance of synthesis-based AD methods, with improvements observed across most metrics.  
This demonstrates that our generation framework can serve as a powerful, plug-and-play data augmentation module to enhance existing anomaly detection methods.

\end{document}